\documentclass[10pt,twocolumn,letterpaper]{article}

\usepackage{cvpr}              
\definecolor{cvprblue}{rgb}{0.21,0.49,0.74}
\usepackage[pagebackref,breaklinks,colorlinks,allcolors=cvprblue]{hyperref}

\usepackage{cuted}
\usepackage{capt-of}
\usepackage{balance}

\def\paperID{11} 
\def\confName{CVPR}
\def\confYear{2026}

\title{PatchPoison: Poisoning Multi-View Datasets to Degrade 3D Reconstruction}

\author{Prajas Wadekar
\quad
Venkata Sai Pranav Bachina\thanks{Equal contribution.\\Accepted at 1st IEEE/CVF CVPR Workshop on Security, Privacy, and Adversarial Robustness in 3D Generative Vision Models (SPAR-3D), 2026.}
\quad
Kunal Bhosikar\footnotemark[1]\\
Ankit Gangwal
\quad
Charu Sharma \\
{\normalsize International Institute of Information Technology, Hyderabad, India}\\
{\tt\small \{prajas.wadekar, kunal.bhosikar\}@research.iiit.ac.in} \quad {\tt\small bachina.pranav@gmail.com} \\ {\tt\small \{gangwal, charu.sharma\}@iiit.ac.in}
}

\begin{document}
\maketitle
\thispagestyle{empty}

\begin{strip}
    \centering
    \includegraphics[width=\linewidth]{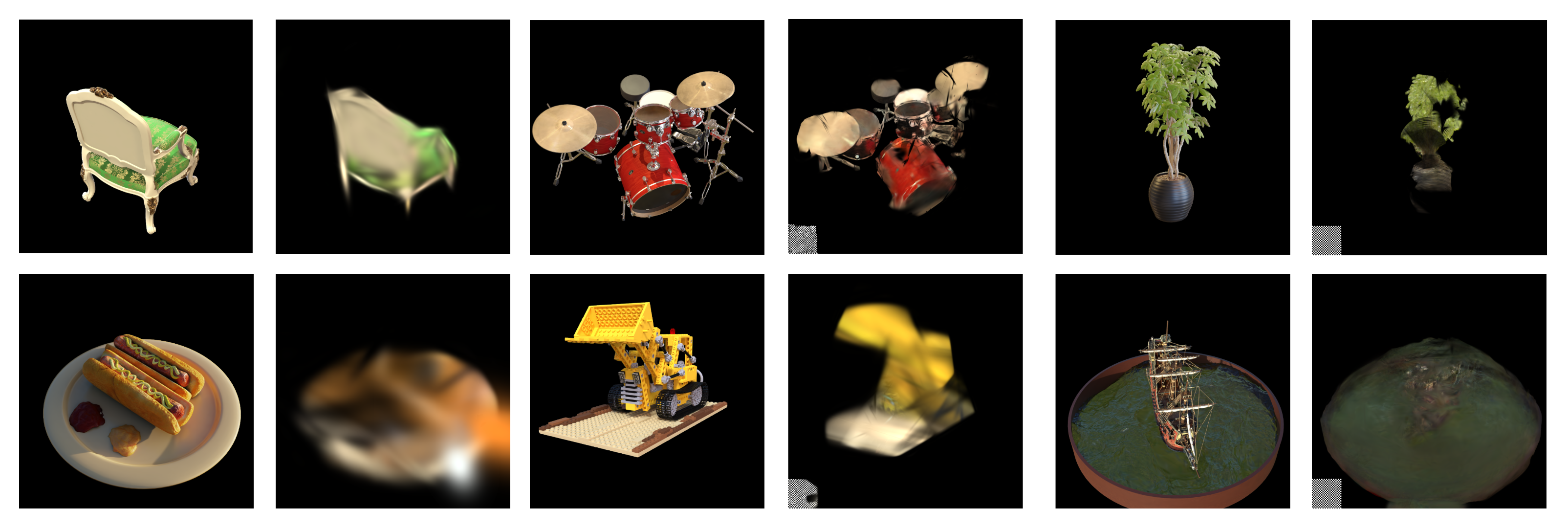}
    \captionof{figure}{Clean vs. poisoned 3DGS reconstructions on the NeRF-Synthetic dataset. Quantitative performance drops significantly due to corrupted feature matching, which reduces valid feature correspondences used during reconstruction.}
    \label{fig:teaser}
\end{strip}

\begin{abstract}
3D Gaussian Splatting (3DGS) has recently enabled highly photorealistic 3D reconstruction from casually captured multi-view images. However, this accessibility raises a privacy concern: publicly available images or videos can be exploited to reconstruct detailed 3D models of scenes or objects without the owner's consent. We present \textbf{PatchPoison}, a lightweight dataset-poisoning method that prevents unauthorized 3D reconstruction. Unlike global perturbations, PatchPoison injects a small high-frequency adversarial patch, a structured checkerboard, into the periphery of each image in a multi-view dataset. The patch is designed to corrupt the feature-matching stage of Structure-from-Motion~(SfM) pipelines such as COLMAP by introducing spurious correspondences that systematically misalign estimated camera poses. Consequently, downstream 3DGS optimization diverges from the correct scene geometry. On the NeRF-Synthetic benchmark, inserting a \(12\times12\) pixel patch increases reconstruction error by \(6.8\times\) in LPIPS, while the poisoned images remain unobtrusive to human viewers. PatchPoison requires no pipeline modifications, offering a practical, ``drop-in" preprocessing step for content creators to protect their multi-view data.
\end{abstract}

\section{Introduction}
\label{sec:intro}

The rapid progress of neural 3D reconstruction has made it increasingly easy to convert a collection of casually captured photographs into a high-fidelity 3D model. Modern approaches such as Neural Radiance Fields (NeRF)~\cite{mildenhall2020nerf} and 3DGS~\cite{kerbl3Dgaussians} can reconstruct photorealistic scenes from a short video or a small set of multi-view images. As a result, a simple video walkthrough of a room, a person, or a physical object can now be used to reconstruct a detailed and renderable 3D representation of the scene. While this capability enables many creative and industrial applications, it also introduces a privacy risk: publicly shared images or videos can be silently repurposed to reconstruct a 3D model of a scene or person without the knowledge or consent of the content owner.

Existing approaches to protecting 3D assets typically operate on protecting the 3DGS model, not the dataset. For example, watermarking and steganography methods~\cite{huang2024gaussianmarker,tan2024water,zhang2024gs} embed ownership signals directly into a trained 3DGS representation, implicitly assuming that reconstruction has already taken place. Other works explore poisoning attacks on the training process itself~\cite{hong2025gausstrap,lu2025poisonsplat}, which require access to or participation in the model training pipeline making them less practical in real-world settings where such access is restricted or unavailable. 
Other Gaussian splatting pipelines like TGC-based geometric cloak \cite{song2024geometry} take in a single image and use Triplane Gaussian splatting to prevent reconstruction using triplane gaussian splatting by making the renders shape a specific way. 
These methods are designed for single-image inputs and do not extend to multi-view settings.
None of these methods protect already shared multi-view images.

Most modern neural reconstruction systems, including NeRF and 3DGS, follow a canonical two-stage pipeline.
First, a SfM stage estimates camera poses and a sparse scene structure by extracting keypoints and matching features across images~\cite{snavely2006photo}. In practice, this step is commonly performed using systems such as COLMAP~\cite{schoenberger2016sfm}. Second, a per-scene optimization stage fits a scene representation, either a neural radiance field or a set of 3D Gaussian primitives, to reproduce the input views given the estimated camera poses. This dependency creates a critical vulnerability: if the camera poses estimated during the SfM stage are incorrect, the downstream reconstruction stage cannot recover a coherent scene representation. Consequently, we identify this dependency as a critical vulnerability that can be leveraged for privacy protection.

In this work, we instead intervene earlier in the pipeline by poisoning the multi-view image dataset~\cite{barron2022mipnerf360} itself. Our key observation is that the feature-matching stage of SfM pipelines such as COLMAP is highly sensitive to repetitive high-frequency patterns. By inserting a small checkerboard patch at a fixed corner of every image in the dataset, we introduce a dense set of artificial keypoints that are consistently detected across views. Because these patch-induced keypoints appear at identical image coordinates in every
view, the feature matcher establishes strong but incorrect correspondences between them. These correspondences are geometrically inconsistent with the true scene structure, causing bundle adjustment to estimate severely distorted camera poses. Once the camera poses are corrupted, downstream neural reconstruction methods such as NeRF or 3DGS fail to recover a coherent scene representation. Importantly, the patch occupies only a tiny portion of each image, making the poisoned dataset visually indistinguishable from the original to human observers.

\smallskip
\noindent\textbf{Contributions.} 
Our contributions are: 
\begin{enumerate}
    \item We identify the feature-matching stage of SfM pipelines as a practical stage for protecting multi-view datasets from unauthorized 3D reconstruction.
    \item We propose PatchPoison, a lightweight dataset-poisoning method that inserts a small checkerboard patch to corrupt camera pose estimation and degrade reconstruction while preserving visual fidelity.
    \item We demonstrate experimentally that a patch as small as \(12\times12\) pixels can significantly degrade 3DGS reconstruction while remaining visually imperceptible in the poisoned images.
\end{enumerate}
\section{Related Works}
\label{sec:related Works}

\paragraph{Neural 3D Reconstruction Pipelines.}
Modern approaches to multi-view 3D reconstruction aim to recover photorealistic scene representations from collections of images. NeRF~\cite{mildenhall2020nerf} represents scenes as continuous volumetric radiance fields parameterized by a multilayer perceptron and render novel views through differentiable volume rendering, while subsequent work such as Mip-NeRF 360~\cite{barron2022mipnerf360} improves reconstruction quality for large and unbounded scenes. More recently, 3DGS~\cite{kerbl3Dgaussians} replaces implicit neural representations with explicit anisotropic 3D Gaussians that can be efficiently rasterized to achieve real-time rendering and high-fidelity reconstruction. Despite their differences in representation, these methods rely on a similar preprocessing pipeline to estimate camera poses and sparse geometry from input images. In practice, SfM systems such as COLMAP~\cite{schoenberger2016sfm}, building upon classical multi-view reconstruction pipelines~\cite{snavely2006photo}, are widely used for camera calibration prior to neural optimization. Recent learning-based alternatives such as DUSt3R~\cite{wang2024dust3r} and
MASt3R~\cite{leroy2024mast3r} attempt to replace traditional feature matching with learned dense correspondences, suggesting potential future directions for extending reconstruction attacks to learning-based pose estimation pipelines.

\paragraph{Adversarial Attacks on 3D Representations.}
Several works have explored adversarial attacks targeting neural 3D representations and rendering systems. Poison-Splat~\cite{lu2025poisonsplat} introduces a poisoning strategy that maximizes Total Variation in reconstructed scenes, causing excessive densification and dramatically increasing the computational cost of 3DGS training. Other work investigates attacks on the behavior of trained models, such as GaussTrap~\cite{hong2025gausstrap}, which implants viewpoint-triggered backdoors into 3DGS models, and IPA-NeRF~\cite{jiang2024ipa}, which optimizes adversarial perturbations over training views to degrade novel-view synthesis. Additional research explores adversarial manipulation of reconstructed 3D models themselves; for example, Gaussian Splatting Under Attack~\cite{zeybey2024gaussian} introduces adversarial perturbations that significantly degrade CLIP-based recognition on rendered views. While these approaches target the training process, rendering behavior, or downstream recognition tasks, they assume access to the reconstruction or training pipeline. In contrast, our work operates purely at the dataset level and aims to corrupt the upstream pose estimation stage used by standard reconstruction pipelines.

\paragraph{Physical and Patch-Based Attacks on Reconstruction.}
Another line of work studies physical adversarial patterns designed to interfere with camera pose estimation during image capture. The Kaleidoscopic Background Attack (KBA)~\cite{ding2025kba} introduces specially designed visual patterns placed in the scene background that disrupt feature matching and degrade reconstruction quality. While conceptually related to our approach, such attacks require printing and physically placing adversarial objects in the scene prior to image capture, and the resulting patterns are visually conspicuous. In contrast, our method operates digitally on already-captured multi-view images and introduces only a small patch occupying a tiny region of the image, making the perturbation unobtrusive to human observers.

\paragraph{Watermarking and Protection of 3D Assets.}
Another direction of research focuses on protecting the ownership of reconstructed 3D assets through watermarking or steganography. Methods such as GS-Hider~\cite{zhang2024gs} and GaussianStego~\cite{li2024gaussianstego} embed hidden messages within trained 3DGS models, while other approaches including WATER-GS~\cite{tan2024water}, GaussianMarker~\cite{huang2024gaussianmarker}, and
GuardSplat~\cite{chen2025guardsplat} insert digital watermarks to enable ownership verification of reconstructed scenes. Geometry Cloak \cite{song2024geometry} integrates watermark in such way that the rendered 3D model from a single image displays a pattern if viewed from a particular view. However, these approaches assume that the 3D reconstruction has already been created. In contrast, our method prevents unauthorized reconstruction by poisoning the multi-view dataset.
\section{Background}
\label{sec:background}

Modern neural 3D reconstruction pipelines typically follow a two-stage process consisting of camera pose estimation from images followed by scene representation optimization. In this section we briefly review the components of this pipeline that are relevant to our technique.

\subsection{Structure-from-Motion (SfM)}
\label{subsec:sfm}

Given a set of images \(\mathcal{I} = \{I_i\}_{i=1}^N\), SfM estimates camera parameters and a sparse scene structure by identifying visual correspondences across images. Systems such as COLMAP~\cite{schoenberger2016sfm} first detect local features (e.g., SIFT keypoints) in each image and match them across image pairs using descriptor similarity. These matches are filtered through geometric verification (typically using RANSAC) to remove inconsistent correspondences. The remaining matches are then used to estimate camera poses and triangulate 3D points. A global bundle adjustment step jointly refines camera poses and 3D points by minimizing reprojection error across all observations.

\subsection{Neural Reconstruction from Known Poses}

Once camera poses have been estimated, neural reconstruction methods optimize a scene representation to reproduce the input views. NeRF~\cite{mildenhall2020nerf} model the scene as a continuous radiance field parameterized by a neural network, while 3DGS~\cite{kerbl3Dgaussians} represents the scene as a collection of anisotropic 3D Gaussians that are efficiently rasterized during rendering. Both approaches assume that camera poses are known and accurate. During training, the scene parameters are optimized so that rendered images from the estimated camera poses match the original input images.

\subsection{Sensitivity to Pose Errors}

The success of neural reconstruction critically depends on the accuracy of the camera poses estimated by SfM. If incorrect feature correspondences are introduced during the matching stage, the resulting bundle adjustment optimization may produce inconsistent or distorted camera poses. Because neural reconstruction methods rely on these poses to establish geometric consistency between views, such pose errors propagate directly to the final reconstruction, often leading to distorted geometry or optimization failure. This makes SfM feature matching a natural attack surface for poisoning.
\section{Methodology}
\label{sec:methodology}

\subsection{Problem Formulation}

Let \(\mathcal{I} = \{I_i\}_{i=1}^{N}\) be a clean multi-view image dataset captured from camera poses $\mathcal{C}_1 = \{C_i\}_{i=1}^{N}$. A Structure-from-Motion pipeline \(\Phi\) estimates camera poses from the dataset: \(\hat{\mathcal{C}}_1 = \Phi(\mathcal{I})\). A reconstruction module \(\mathcal{R}\) then produces a scene representation \(G_1 = \mathcal{R}(\mathcal{I}, \hat{\mathcal{C}}_1)\).

Our goal is to construct a poisoned dataset \(\mathcal{I}^* = \{I_i^*\}_{i=1}^N\) such that:
\begin{align}
    \quad & D(I_i, I_i^*) \approx 0 \quad \forall i
    \label{eq:imperceptibility} \\
    \quad &
    D\big(
    \mathcal{R}(\mathcal{I}^*, \Phi(\mathcal{I}^*)), 
    G_1) \gg \epsilon
    \label{eq:degradation}
\end{align}
\noindent
where Eq.~\eqref{eq:imperceptibility} enforces \textit{imperceptibility} of the perturbation, and Eq.~\eqref{eq:degradation} enforces \textit{reconstruction degradation}. The distance function \(D(\cdot)\) measures perceptual similarity using metrics such as SSIM, PSNR, and LPIPS.

\subsection{Pipeline}

\paragraph{Targeting the SfM stage.}
COLMAP's reconstruction pipeline consists of feature extraction (e.g., SIFT), feature matching, geometric verification (RANSAC), and bundle adjustment. Among these stages, feature matching is particularly vulnerable: it relies on local appearance descriptors to establish correspondences between images. Patterns that generate strong, repeatable high-frequency responses can produce consistent but incorrect matches across views.

\begin{figure*}
    \centering
    \includegraphics[width=0.78\linewidth]{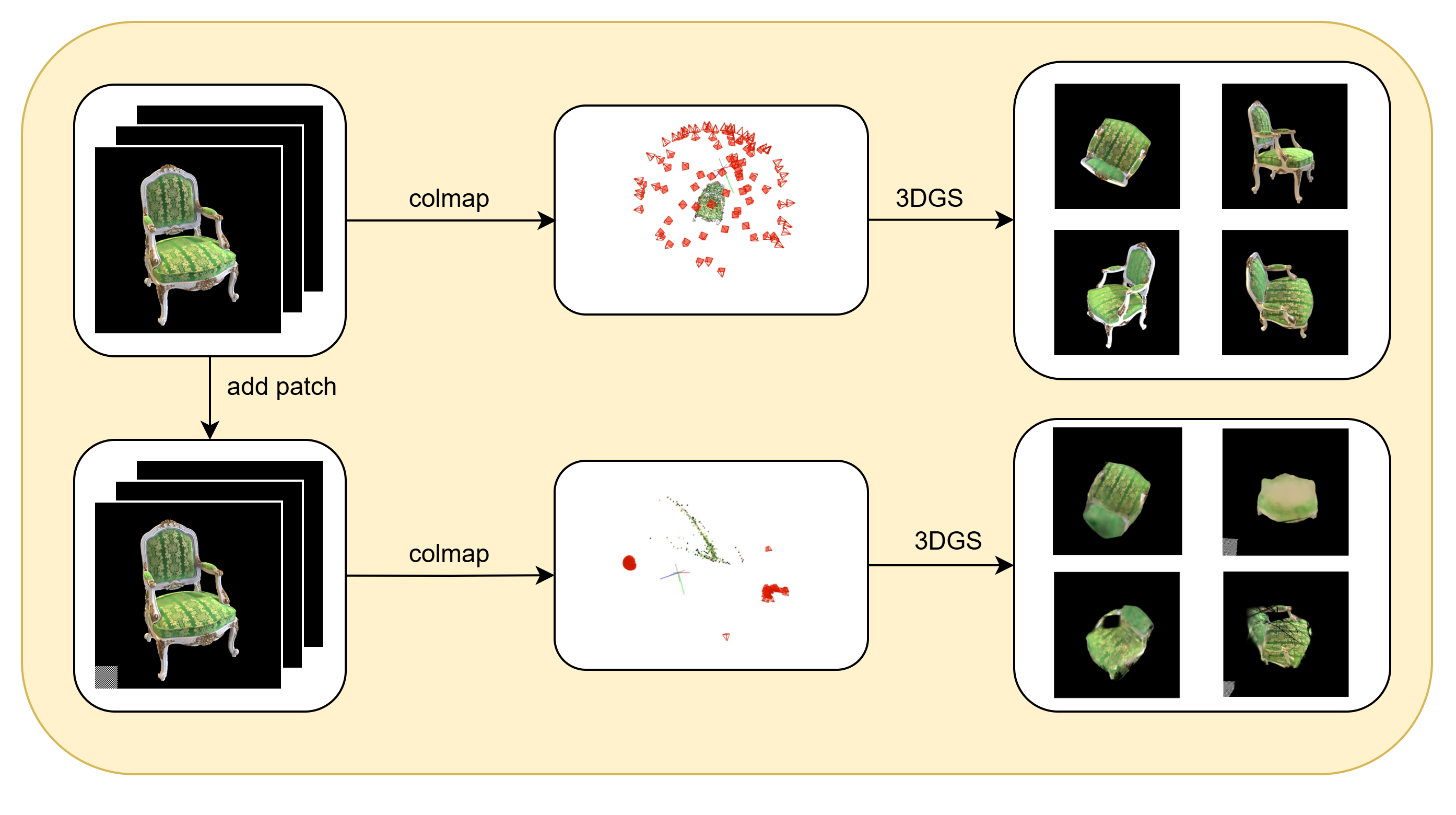}
    \caption{\textbf{PatchPoison pipeline.} A small high-frequency patch inserted into each image induces spurious feature correspondences during Structure-from-Motion, leading to incorrect camera pose estimation and degraded 3D Gaussian Splatting reconstruction.}
    \label{fig:pipeline}
\end{figure*}

\paragraph{Patch-induced spurious correspondences.}
A checkerboard pattern contains dense high-frequency structures that produce numerous corner-like keypoints with strong descriptor responses. When the same patch is inserted at a fixed image location across all views, these patch-induced keypoints appear consistent across images and are matched with high confidence by the SfM pipeline. However, these correspondences are not associated with any real 3D structure in the scene and therefore violate geometric consistency constraints (e.g., epipolar geometry). As a result, bundle adjustment attempts to reconcile incompatible constraints, leading to incorrect or unstable camera pose estimates. These corrupted poses subsequently cause downstream neural reconstruction methods such as 3D Gaussian Splatting to fail or produce distorted geometry (Fig.~\ref{fig:pipeline}).

\paragraph{Patch construction.}
Given an image \(I_i \in \mathbb{R}^{H \times W \times 3}\), we generate a checkerboard patch \(M \in \{0,1\}^{P \times P}\) with block size \(b\) pixels. The patch is placed at a fixed corner of the image and blended with strength \(\alpha \in [0,1]\). The poisoned image is defined as:
\begin{equation}
    I_i^* = I_i \odot (1 - \alpha M_\Omega) + \alpha M_\Omega,
\end{equation}
where \(M_\Omega\) denotes the patch embedded in region \(\Omega\) and zero elsewhere. The patch is deterministic and parameterized by \((P, b, \alpha, \Delta c)\). In practice, we choose parameters such that the patch occupies a small fraction of the image while still inducing strong feature responses.

\subsection{Evaluation Protocol}
\label{sec:eval_protocol}

We evaluate the effectiveness of the attack using two complementary criteria.

\paragraph{Reconstruction degradation \((\text{cost}_1)\).}
Let \(G_2 = \mathcal{R}(\mathcal{I}^*, \Phi(\mathcal{I}^*))\) be the reconstructed scene from the poisoned dataset, and let \(\mathcal{C}_2 = \Phi(\mathcal{I}^*)\) be the corresponding camera poses. We measure reconstruction quality by comparing rendered images against the poisoned inputs:
\begin{equation}
  \text{cost}_1 = \bigl\{\operatorname{SSIM}(\mathcal{I}^*, \mathcal{R}(G_2, \mathcal{C}_2)),\,
                         \operatorname{PSNR}(\cdot),\,
                         \operatorname{LPIPS}(\cdot)\bigr\}
  \label{eq:cost1}
\end{equation}
which should be \emph{low}, indicating poor reconstruction quality.

\paragraph{Imperceptibility \((\text{cost}_2)\).}
We measure visual similarity between clean and poisoned datasets as:
\begin{equation}
  \text{cost}_2 = \bigl\{\operatorname{SSIM}(\mathcal{I}, \mathcal{I}^*),\,
                         \operatorname{PSNR}(\cdot),\,
                         \operatorname{LPIPS}(\cdot)\bigr\}
  \label{eq:cost2}
\end{equation}
which should be \emph{high}, ensuring that perturbations remain visually imperceptible.
\section{Experiments}
\label{sec:experiments}

\subsection{Setup}

\paragraph{Datasets.}
We evaluate PatchPoison on the \textit{NeRF-Synthetic}~\cite{mildenhall2020nerf} dataset and the \textit{Mip-NeRF 360}~\cite{barron2022mipnerf360} dataset, covering both object-centric scenes and complex real-world unbounded environments.

\paragraph{Reconstruction pipeline.}
We use the official 3D Gaussian Splatting implementation~\cite{kerbl3Dgaussians} with COLMAP for camera pose estimation. All models are trained for 30,000 iterations using default hyperparameters.

\paragraph{Baselines.}
As no prior work studies dataset poisoning for multi-view reconstruction, we compare PatchPoison against a range of standard baselines: (i) a \textit{clean dataset}, representing an upper bound on
reconstruction quality; (ii) \textit{geometric transformations} of the input images, including rotations, scaling, shearing, and translational shifts; and (iii) \textit{image perturbations} such as Gaussian blur, Gaussian noise, and JPEG compression.

\paragraph{Metrics.}
We evaluate reconstruction quality by comparing rendered images from the reconstructed scene against the poisoned inputs (\textit{Poisoned vs Render}), and imperceptibility by comparing poisoned images against the original inputs (\textit{Poisoned vs Original}). We report SSIM, PSNR, and LPIPS, averaged over three independent runs.

\subsection{Ablations}

We systematically analyze the design choices of PatchPoison to understand which factors contribute most to reconstruction degradation while preserving imperceptibility.

\paragraph{Patch size.} 
We vary the patch size \(P \in \{12, 24, 48, 76, 100, 148, 200\}\) pixels on \(800\times800\) images, corresponding to approximately \(1.5\%\) to \(25\%\) of the image width.

\paragraph{Spatial frequency.} 
We control the frequency of the checkerboard pattern via the block size \(b \in \{1, 2, 4, 8, 10, 16, 20, 25, 50\}\) pixels.

\paragraph{Color contrast.} 
We vary the intensity difference between bright and dark regions of the checkerboard using \(\Delta c \in \{5, 10, 25, 50, 75, 100, 125, 150, 175, 200\}\).

\paragraph{Alpha blending.} 
We control patch visibility using blending strength \(\alpha \in \{5, 10, \ldots, 255\}/255\).

\paragraph{Pattern type.} 
We evaluate different pattern structures, including checkerboards, lines, circles, and their combinations, to analyze how their geometric layout and spatial frequency affect feature detection and matching.

\paragraph{Poisoning ratio.} 
We vary the fraction of poisoned training images from \(5\%\) to \(100\%\) to evaluate the robustness of the method under partial poisoning.
\section{Results}
\label{sec:results}

For all experiments, we evaluate our approach using two primary criteria: reconstruction quality and imperceptibility. Reconstruction quality (\textit{Poisoned vs. Render}) measures the fidelity of the model's output, computed as the reconstruction cost between the rendered images and the poisoned training dataset containing the high-frequency patch. Imperceptibility (\textit{Poisoned vs. Original}) quantifies the stealthiness of the modification by calculating the difference between the poisoned images and the clean, original dataset. Results on the NeRF-Synthetic dataset are averaged across all eight scenes. In cases where the poisoned and original images remain identical, the Peak Signal-to-Noise Ratio (PSNR) is infinite.


\paragraph{Overview.}
PatchPoison consistently degrades reconstruction quality while preserving visual fidelity of the input images. Even small perturbations significantly disrupt the reconstruction pipeline by corrupting feature correspondences, while remaining nearly imperceptible.

\begin{figure}
    \centering
    \includegraphics[width=\linewidth]{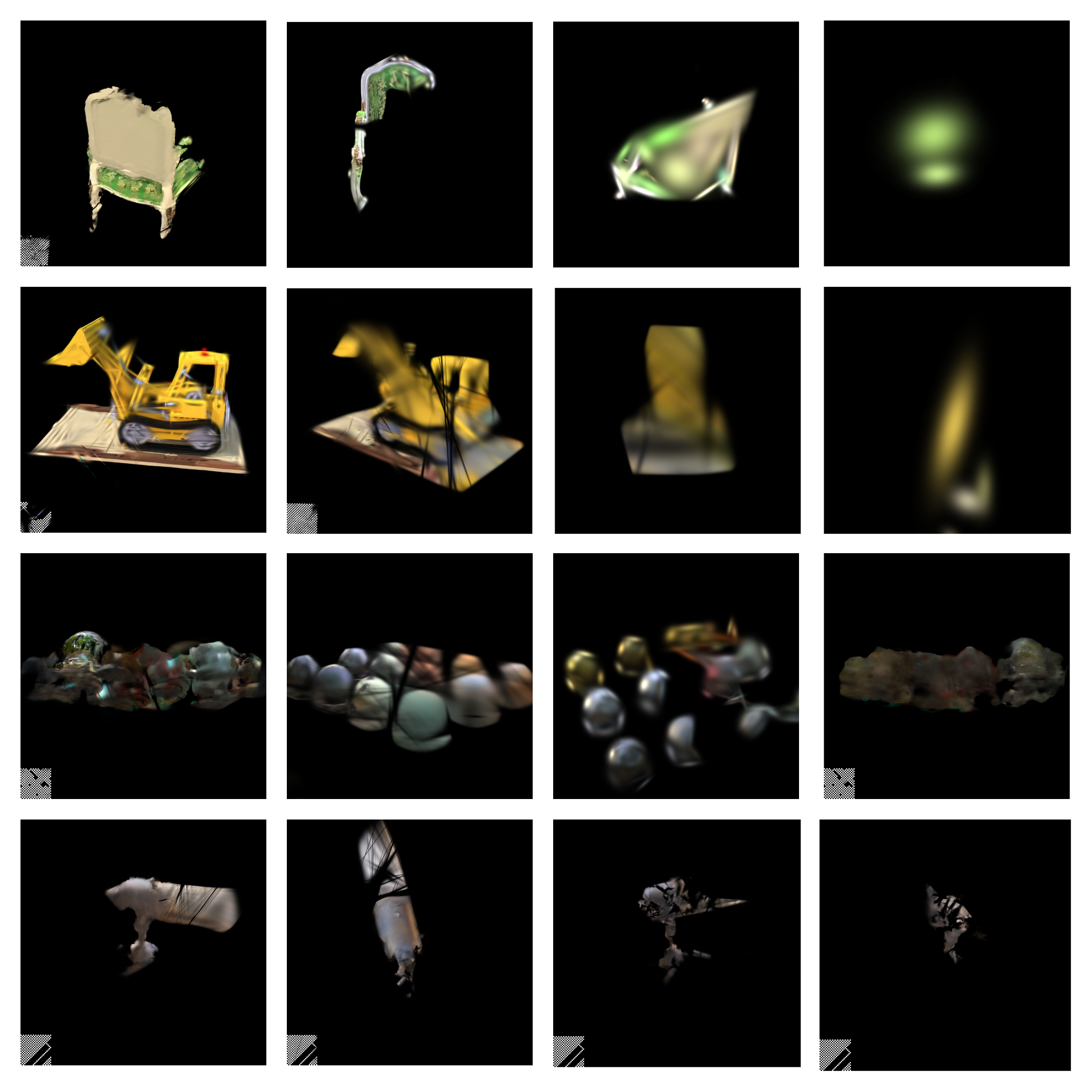}
    \caption{\textbf{Qualitative evaluation on the NeRF-Synthetic dataset.} These renders are generated from the poisoned images, illustrating the structural degradation and severe visual artifacts introduced by the PatchPoison attack across multiple views. We omit views that resulted in entirely black renders, which constituted the majority of the overall outputs.}
    \label{fig:reconstruction}
\end{figure}

\paragraph{Why checkerboard patterns are effective.}
SIFT detects scale-space extrema in the Difference-of-Gaussian pyramid. A checkerboard pattern with block size \(b=4\)--\(8\) pixels generates dense high-contrast extrema at the native image scale, producing distinctive descriptors. Since the patch is placed at identical pixel locations across all views, these descriptors match with very high confidence, dominating the matching stage. These correspondences are geometrically inconsistent, causing bundle adjustment to fail and yielding distorted reconstructions (Fig.~\ref{fig:reconstruction}).

\subsection{Effect of Patch Size}
Table~\ref{tab:patch_results} shows the effect of patch size on reconstruction degradation and imperceptibility. Even very small patches (\(12\times12\) pixels) significantly degrade reconstruction quality (SSIM drops from \(0.969\) to \(0.793\)). As patch size increases, degradation initially improves; however, for very large patches (e.g., \(200\) pixels), performance partially recovers because the model begins to reconstruct the checkerboard pattern itself.

\begin{table}
    \centering
    \resizebox{\columnwidth}{!}{
    \begin{tabular}{lcccccc}
        \toprule
        Patch Size & \multicolumn{3}{c}{Poisoned vs Render} & \multicolumn{3}{c}{Poisoned vs Original} \\
        \cmidrule(lr){2-4} \cmidrule(lr){5-7}
         & SSIM$\downarrow$ & PSNR$\downarrow$ & LPIPS$\uparrow$ & SSIM$\uparrow$ & PSNR$\uparrow$ & LPIPS$\downarrow$ \\
        \midrule
        clean  & 0.969 $\pm$ 0.026 & 32.96 $\pm$ 3.26 & 0.050 $\pm$ 0.035 & 1.000 & inf & 0.000 \\
        \textbf{12}  & 0.793 ± 0.106 & 18.06 ± 5.82 & 0.336 ± 0.147 & 0.9996 & 39.03 & 0.0016 \\
        24  & 0.798 ± 0.110 & 18.32 ± 5.82 & 0.322 ± 0.155 & 0.9988 & 33.47 & 0.0036 \\
        48  & 0.775 ± 0.091 & 16.91 ± 3.06 & 0.354 ± 0.121 & 0.9958 & 27.45 & 0.0083 \\
        76  & 0.809 ± 0.101 & 17.42 ± 4.80 & 0.302 ± 0.165 & 0.9900 & 23.44 & 0.0163 \\
        100 & 0.773 ± 0.072 & 16.19 ± 2.64 & 0.365 ± 0.096 & 0.9831 & 21.07 & 0.0259 \\
        148 & 0.770 ± 0.075 & 15.75 ± 3.25 & 0.374 ± 0.106 & 0.9639 & 17.69 & 0.0468 \\
        200 & 0.774 ± 0.117 & 16.32 ± 6.51 & 0.348 ± 0.163 & 0.9350 & 15.14 & 0.0738 \\
        \bottomrule
    \end{tabular}
    }
    \caption{Reconstruction Degradation and Imperceptibility for different scenes in the NeRF-Synthetic dataset with different sizes of adversarial patch.}
    \label{tab:patch_results}
\end{table}

Imperceptibility decreases monotonically with patch size, highlighting that smaller patches provide the best trade-off between effectiveness and stealth.

Table~\ref{tab:per_scene_breakdown} shows consistent degradation across all scenes, demonstrating robustness of the attack.

\begin{table}
    \centering
    \resizebox{\columnwidth}{!}{
    \begin{tabular}{l ccc ccc}
        \toprule
        & \multicolumn{3}{c}{Poisoned vs Render} & \multicolumn{3}{c}{Poisoned vs Original} \\
        \cmidrule(lr){2-4} \cmidrule(lr){5-7}
        Scene & SSIM$\uparrow$ & PSNR$\uparrow$ & LPIPS$\downarrow$ & SSIM$\downarrow$ & PSNR$\downarrow$ & LPIPS$\uparrow$ \\
        \midrule
        \multicolumn{7}{c}{\textbf{Clean Baseline}} \\
        \midrule
        chair     & 0.990 $\pm$ 0.000 & 33.54 $\pm$ 0.00 & 0.023 $\pm$ 0.000 & 1.000 & inf & 0.000 \\
        drums     & 0.966 $\pm$ 0.000 & 27.91 $\pm$ 0.00 & 0.057 $\pm$ 0.000 & 1.000 & inf & 0.000 \\
        ficus     & 0.993 $\pm$ 0.000 & 35.90 $\pm$ 0.10 & 0.010 $\pm$ 0.000 & 1.000 & inf & 0.000 \\
        hotdog    & 0.989 $\pm$ 0.000 & 37.58 $\pm$ 0.01 & 0.025 $\pm$ 0.000 & 1.000 & inf & 0.000 \\
        lego      & 0.965 $\pm$ 0.001 & 32.72 $\pm$ 0.02 & 0.051 $\pm$ 0.001 & 1.000 & inf & 0.000 \\
        materials & 0.957 $\pm$ 0.000 & 29.64 $\pm$ 0.02 & 0.076 $\pm$ 0.000 & 1.000 & inf & 0.000 \\
        mic       & 0.981 $\pm$ 0.000 & 35.00 $\pm$ 0.09 & 0.036 $\pm$ 0.001 & 1.000 & inf & 0.000 \\
        ship      & 0.912 $\pm$ 0.000 & 31.42 $\pm$ 0.00 & 0.124 $\pm$ 0.000 & 1.000 & inf & 0.000 \\
        \midrule
        \multicolumn{7}{c}{\textbf{Poisoned (size 100 , block size 4)}} \\
        \midrule
        chair     & 0.798 $\pm$ 0.002 & 13.19 $\pm$ 0.07 & 0.346 $\pm$ 0.001 & 0.983 & 21.07 & 0.027 \\
        drums     & 0.799 $\pm$ 0.001 & 15.40 $\pm$ 0.10 & 0.359 $\pm$ 0.003 & 0.983 & 21.07 & 0.026 \\
        ficus     & 0.862 $\pm$ 0.001 & 18.46 $\pm$ 0.08 & 0.185 $\pm$ 0.001 & 0.983 & 21.07 & 0.027 \\
        hotdog    & 0.735 $\pm$ 0.008 & 15.28 $\pm$ 0.08 & 0.424 $\pm$ 0.001 & 0.983 & 21.07 & 0.026 \\
        lego      & 0.679 $\pm$ 0.001 & 12.49 $\pm$ 0.19 & 0.485 $\pm$ 0.007 & 0.983 & 21.07 & 0.026 \\
        materials & 0.752 $\pm$ 0.005 & 16.13 $\pm$ 0.15 & 0.384 $\pm$ 0.005 & 0.983 & 21.07 & 0.027 \\
        mic       & 0.872 $\pm$ 0.004 & 18.95 $\pm$ 0.39 & 0.288 $\pm$ 0.001 & 0.983 & 21.07 & 0.026 \\
        ship      & 0.690 $\pm$ 0.009 & 19.63 $\pm$ 0.18 & 0.452 $\pm$ 0.005 & 0.983 & 21.07 & 0.021 \\
        \bottomrule
    \end{tabular}
    }
    \caption{Per-scene reconstruction and imperceptibility metrics on NeRF-Synthetic, comparing clean and poisoned data (patch size 100) over 3 runs.}
    \label{tab:per_scene_breakdown}
\end{table}


\subsection{Effect of Spatial Frequency}
Table~\ref{tab:freq_results} evaluates the effect of block size. Very small blocks \((b=1,2)\) are ineffective because the pattern resembles noise and is often ignored by feature detectors. Moderate frequencies \((b=4)\) produce the strongest degradation, as they generate stable, repeatable keypoints that dominate feature matching and introduce spurious correspondences.

\begin{table}
    \centering
    \resizebox{\columnwidth}{!}{
    \begin{tabular}{lcccccc}
        \toprule
        Block Size & \multicolumn{3}{c}{Poisoned vs Render} & \multicolumn{3}{c}{Poisoned vs Original} \\
        \cmidrule(lr){2-4} \cmidrule(lr){5-7}
         & SSIM$\downarrow$ & PSNR$\downarrow$ & LPIPS$\uparrow$ & SSIM$\uparrow$ & PSNR$\uparrow$ & LPIPS$\downarrow$ \\
        \midrule
        clean  & 0.969 $\pm$ 0.026 & 32.97 $\pm$ 3.26 & 0.050 $\pm$ 0.035 & 1.000 & inf & 0.000 \\
        1  & 0.952 $\pm$ 0.029 & 20.63 $\pm$ 0.34 & 0.073 $\pm$ 0.038 & 0.983 & 21.07 & 0.022 \\
        2  & 0.953 $\pm$ 0.027 & 20.62 $\pm$ 0.32 & 0.080 $\pm$ 0.035 & 0.983 & 21.07 & 0.031 \\
        \textbf{4}  & \textbf{0.788} $\pm$ \textbf{0.106} & \textbf{16.75} $\pm$ \textbf{4.54} & \textbf{0.342} $\pm$ \textbf{0.154} & \textbf{0.983} & \textbf{21.07} & \textbf{0.026} \\
        6  & 0.751 $\pm$ 0.068 & 14.66 $\pm$ 3.14 & 0.398 $\pm$ 0.096 & 0.983 & 21.07 & 0.025 \\
        8  & 0.828 $\pm$ 0.131 & 17.75 $\pm$ 3.16 & 0.268 $\pm$ 0.195 & 0.983 & 21.07 & 0.026 \\
        10 & 0.795 $\pm$ 0.065 & 16.80 $\pm$ 3.48 & 0.348 $\pm$ 0.096 & 0.984 & 21.07 & 0.028 \\
        16 & 0.850 $\pm$ 0.071 & 18.80 $\pm$ 1.75 & 0.244 $\pm$ 0.116 & 0.985 & 21.07 & 0.028 \\
        25 & 0.842 $\pm$ 0.103 & 18.30 $\pm$ 3.03 & 0.251 $\pm$ 0.146 & 0.988 & 21.07 & 0.026 \\
        50 & 0.815 $\pm$ 0.150 & 17.98 $\pm$ 5.66 & 0.309 $\pm$ 0.232 & 0.990 & 21.07 & 0.022 \\
        \bottomrule
    \end{tabular}
    }
    \caption{Reconstruction degradation and imperceptibility across different checkerboard block sizes (frequency of the pattern) for NeRF-Synthetic Dataset.}
    \label{tab:freq_results}
\end{table}


\subsection{Effect of Pattern Type}

\begin{figure}
    \centering
    \includegraphics[width=\linewidth]{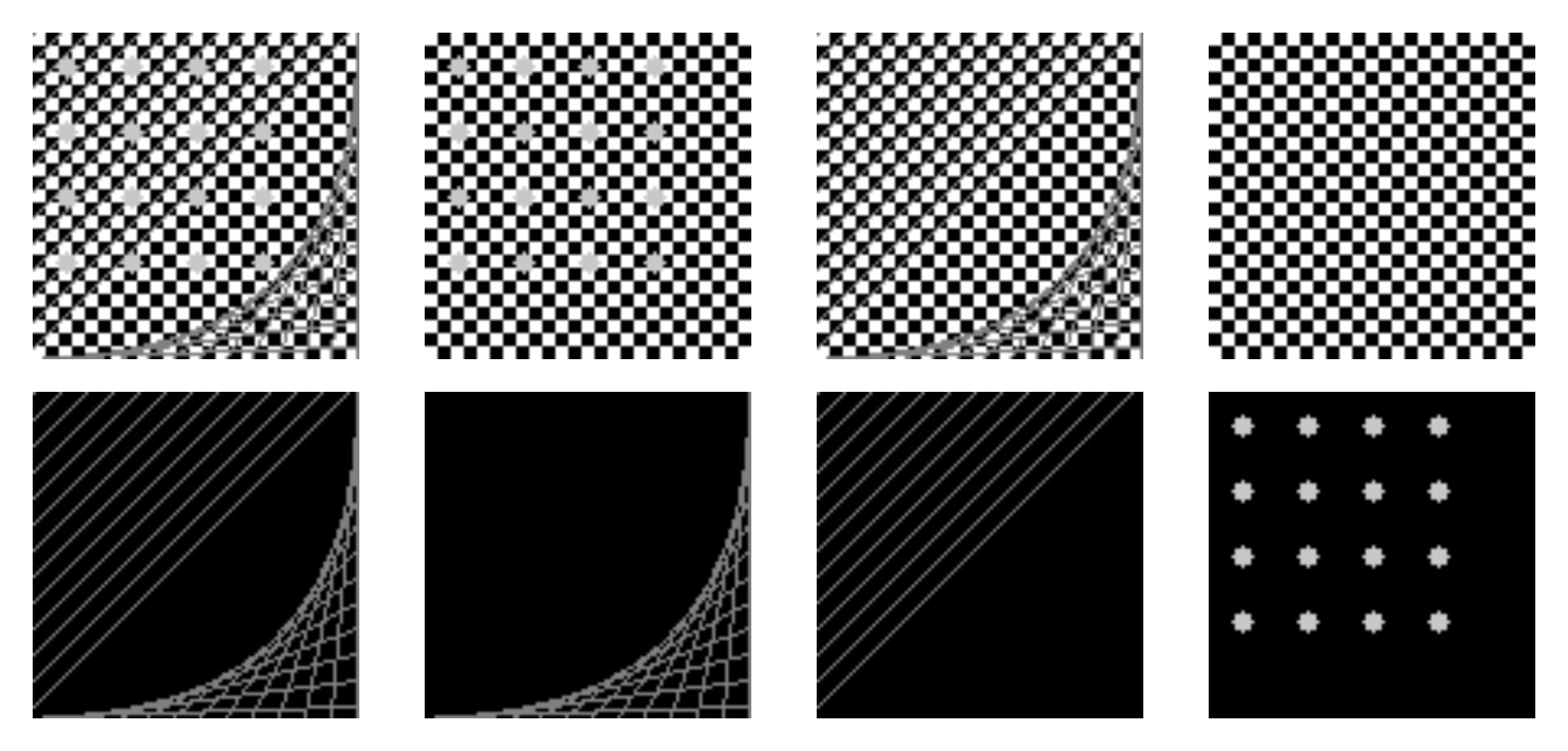}
    \caption{Examples of synthetic pattern variants evaluated.}
    \label{fig:patterns}
\end{figure}

Table~\ref{tab:ns_variants_results} compares different pattern structures. Checkerboard patterns achieve the strongest degradation, confirming that dense corner-like structures are most effective at generating consistent but incorrect correspondences. Other patterns such as lines or circles are less effective due to weaker feature responses. Figure \ref{fig:patterns} shows how the different patches look like. 

\begin{table}
    \centering
    \resizebox{\columnwidth}{!}{
    \begin{tabular}{lcccccc}
        \toprule
        Variant & \multicolumn{3}{c}{Poisoned vs Render} & \multicolumn{3}{c}{Poisoned vs Original} \\
        \cmidrule(lr){2-4} \cmidrule(lr){5-7}
          & SSIM$\uparrow$ & PSNR$\uparrow$ & LPIPS$\downarrow$ & SSIM$\downarrow$ & PSNR$\downarrow$ & LPIPS$\uparrow$ \\
        \midrule
        All Patterns & 0.822 $\pm$ 0.053 & 18.97 $\pm$ 1.96 & 0.294 $\pm$ 0.074 & 0.983 & 21.23 & 0.026 \\
        Checkerboard + Circles & 0.809 $\pm$ 0.088 & 17.98 $\pm$ 3.10 & 0.304 $\pm$ 0.132 & 0.983 & 21.02 & 0.027 \\
        Checkerboard + Diagonal Lines & 0.820 $\pm$ 0.078 & 18.49 $\pm$ 1.82 & 0.291 $\pm$ 0.113 & 0.983 & 21.29 & 0.026 \\
        \textbf{Checkerboard Only} & \textbf{0.801} $\pm$ \textbf{0.094} & \textbf{16.91} $\pm$ \textbf{3.53} & \textbf{0.324} $\pm$ \textbf{0.146} & \textbf{0.983} & \textbf{21.07} & \textbf{0.026} \\
        Circles Only & 0.878 $\pm$ 0.115 & 23.89 $\pm$ 6.66 & 0.190 $\pm$ 0.155 & 0.996 & 33.50 & 0.016 \\
        Diagonal Lines + Circles & 0.859 $\pm$ 0.095 & 21.24 $\pm$ 4.68 & 0.230 $\pm$ 0.143 & 0.986 & 30.00 & 0.023 \\
        Diagonal Lines Only & 0.773 $\pm$ 0.096 & 17.47 $\pm$ 4.81 & 0.353 $\pm$ 0.133 & 0.987 & 32.45 & 0.022 \\
        Intersecting Lines & 0.809 $\pm$ 0.102 & 19.00 $\pm$ 6.20 & 0.314 $\pm$ 0.144 & 0.996 & 34.97 & 0.016 \\
        Parallel Lines & 0.956 $\pm$ 0.032 & 30.33 $\pm$ 2.93 & 0.073 $\pm$ 0.046 & 0.991 & 36.00 & 0.015 \\
\bottomrule
    \end{tabular}
    }
    \caption{
    Reconstruction quality and imperceptibility across different synthetic patterns variants on NeRF-Synthetic scenes.}
     \label{tab:ns_variants_results}
\end{table}

\subsection{Effect of Color Contrast and Transparency}
Table~\ref{tab:ns_alpha_across_scene} shows that even low transparency \((\alpha = 0.1)\) is sufficient to significantly degrade reconstruction while maintaining high imperceptibility. Increasing \(\alpha\) provides limited additional degradation but reduces visual similarity.

\begin{table}
    \centering
    \resizebox{\columnwidth}{!}{
    \begin{tabular}{lcccccc}
        \toprule
        Alpha & \multicolumn{3}{c}{Poisoned vs Render} & \multicolumn{3}{c}{Poisoned vs Original} \\
        \cmidrule(lr){2-4} \cmidrule(lr){5-7}
         & SSIM$\downarrow$ & PSNR$\downarrow$ & LPIPS$\uparrow$ & SSIM$\uparrow$ & PSNR$\uparrow$ & LPIPS$\downarrow$ \\
        \midrule
        clean  & 0.969 $\pm$ 0.026 & 32.97 $\pm$ 3.26 & 0.050 $\pm$ 0.035 & 1.000 & inf & 0.000 \\
        0.001 & 0.970 $\pm$ 0.029 & 33.26 $\pm$ 3.67 & 0.050 $\pm$ 0.039 & 1.000 & inf & 0.000 \\
        \textbf{0.1}   & \textbf{0.783} $\pm$ \textbf{0.073} & \textbf{17.49} $\pm$ \textbf{2.48} & \textbf{0.349} $\pm$ \textbf{0.096} & \textbf{0.984} & \textbf{40.89} & \textbf{0.015} \\
        0.25  & 0.774 $\pm$ 0.070 & 16.57 $\pm$ 3.49 & 0.364 $\pm$ 0.097 & 0.984 & 33.07 & 0.019 \\
        0.5   & 0.770 $\pm$ 0.064 & 16.37 $\pm$ 3.47 & 0.374 $\pm$ 0.088 & 0.983 & 27.05 & 0.023 \\
        0.75  & 0.787 $\pm$ 0.089 & 16.95 $\pm$ 2.88 & 0.344 $\pm$ 0.133 & 0.983 & 23.58 & 0.025 \\
        1     & 0.803 $\pm$ 0.084 & 17.05 $\pm$ 4.60 & 0.321 $\pm$ 0.138 & 0.983 & 21.07 & 0.026 \\
        \bottomrule
    \end{tabular}
    }
    \caption{
    Reconstruction quality and imperceptibility across different transparency value for the data on NeRF-Synthetic scenes, reported as mean $\pm$ std across scenes.}
    \label{tab:ns_alpha_across_scene}
\end{table}

Table~\ref{tab:ns_color_across_scene} shows that low contrast values fail to disrupt reconstruction, while moderate contrast \((\Delta c = 25)\) achieves a good balance. Higher contrast improves attack consistency across scenes but reduces imperceptibility.

\begin{table}
    \centering
    \resizebox{\columnwidth}{!}{
    \begin{tabular}{lcccccc}
        \toprule
        White & \multicolumn{3}{c}{Poisoned vs Render} & \multicolumn{3}{c}{Poisoned vs Original} \\
        \cmidrule(lr){2-4} \cmidrule(lr){5-7}
         & SSIM$\downarrow$ & PSNR$\downarrow$ & LPIPS$\uparrow$ & SSIM$\uparrow$ & PSNR$\uparrow$ & LPIPS$\downarrow$ \\
        \midrule
        clean  & 0.969 $\pm$ 0.026 & 32.97 $\pm$ 3.26 & 0.050 $\pm$ 0.035 & 1.000 & inf & 0.000 \\ 
        5   & 0.964 $\pm$ 0.028 & 33.41 $\pm$ 3.54 & 0.050 $\pm$ 0.038 & 0.992 & 55.11 & 0.003 \\
        10  & 0.958 $\pm$ 0.029 & 33.43 $\pm$ 4.22 & 0.056 $\pm$ 0.040 & 0.987 & 49.15 & 0.008 \\
        \textbf{25}  & \textbf{0.795} $\pm$ \textbf{0.104} & \textbf{18.62} $\pm$ \textbf{5.28} & \textbf{0.331} $\pm$ \textbf{0.152} & \textbf{0.984} & \textbf{41.23} & \textbf{0.014} \\
        50  & 0.781 $\pm$ 0.073 & 17.18 $\pm$ 3.54 & 0.357 $\pm$ 0.102 & 0.984 & 35.22 & 0.018 \\
        100 & 0.805 $\pm$ 0.080 & 18.80 $\pm$ 2.91 & 0.310 $\pm$ 0.120 & 0.983 & 29.20 & 0.021 \\
        200 & 0.781 $\pm$ 0.068 & 16.58 $\pm$ 3.29 & 0.360 $\pm$ 0.094 & 0.983 & 23.18 & 0.025 \\
        255 & 0.772 $\pm$ 0.080 & 16.18 $\pm$ 3.30 & 0.366 $\pm$ 0.106 & 0.983 & 21.07 & 0.026 \\
        \bottomrule
    \end{tabular}
    }
    \caption{Reconstruction quality and imperceptibility across different white values which represents the color difference between the checkerboard pattern for NeRF-Synthetic scenes.}
    \label{tab:ns_color_across_scene}
\end{table}


\subsection{Effect of Training Data Volume}
Table~\ref{tab:ns_limited_training} shows that PatchPoison remains effective even when only \(50\%\) of the dataset is poisoned. At lower poisoning rates (e.g., \(25\%\)), the attack shows higher variance, indicating inconsistent degradation.

\begin{table}
    \centering
    \resizebox{\columnwidth}{!}{
    \begin{tabular}{lcccccc}
        \toprule
        Poison (\%) & \multicolumn{3}{c}{Poisoned vs Render} & \multicolumn{3}{c}{Poisoned vs Original} \\
        \cmidrule(lr){2-4} \cmidrule(lr){5-7}
         & SSIM$\downarrow$ & PSNR$\downarrow$ & LPIPS$\uparrow$ & SSIM$\uparrow$ & PSNR$\uparrow$ & LPIPS$\downarrow$ \\
        \midrule
        clean  & 0.969 $\pm$ 0.026 & 32.97 $\pm$ 3.26 & 0.050 $\pm$ 0.035 & 1.000 & inf & 0.000 \\
        5   & 0.967 $\pm$ 0.028 & 32.08 $\pm$ 3.15 & 0.054 $\pm$ 0.038 & 0.9992 & inf & 0.0013 \\
        10  & 0.937 $\pm$ 0.095 & 29.68 $\pm$ 7.85 & 0.106 $\pm$ 0.167 & 0.9983 & inf & 0.0026 \\
        25  & 0.857 $\pm$ 0.138 & 22.47 $\pm$ 6.11 & 0.222 $\pm$ 0.198 & 0.9958 & inf & 0.0065 \\
        \textbf{50}  & \textbf{0.762} $\pm$ \textbf{0.082} & \textbf{15.16} $\pm$ \textbf{3.68} & \textbf{0.386} $\pm$ \textbf{0.096} & \textbf{0.9915} & \textbf{inf} & \textbf{0.0129} \\
        75  & 0.782 $\pm$ 0.113 & 16.54 $\pm$ 5.91 & 0.352 $\pm$ 0.168 & 0.9873 & inf & 0.0194 \\
        100 & 0.761 $\pm$ 0.084 & 15.68 $\pm$ 3.43 & 0.378 $\pm$ 0.102 & 0.9831 & 21.07 & 0.0258 \\
        \bottomrule
    \end{tabular}
    }
    \caption{Reconstruction quality and imperceptibility under varying poisoning levels with limited training data on NeRF-Synthetic scenes, reported as mean $\pm$ std across scenes.}
    \label{tab:ns_limited_training}
\end{table}


\subsection{Comparison with Baselines}
We compare PatchPoison against Gaussian blur, Gaussian noise, geometric transformations, and JPEG compression.

Gaussian blur (Table~\ref{tab:ns_gaussian_blur_results}) preserves reconstruction quality, indicating that smoothing does not disrupt feature matching. Table~\ref{tab:ns_gaussian_noise_results} shows that gaussian noise degrades reconstruction but significantly reduces imperceptibility, making it visually noticeable. Geometric transformations (Table~\ref{tab:ns_geometric_results}) do not significantly affect reconstruction, as SfM remains robust to such transformations. JPEG compression (Table~\ref{tab:ns_jpeg_compression_results}) produces only marginal degradation.

\begin{table}
    \centering
    \resizebox{\columnwidth}{!}{
    \begin{tabular}{lcccccc}
        \toprule
        Kernel Size & \multicolumn{3}{c}{Poisoned vs Render} & \multicolumn{3}{c}{Poisoned vs Original} \\
        \cmidrule(lr){2-4} \cmidrule(lr){5-7}
         & SSIM$\uparrow$ & PSNR$\uparrow$ & LPIPS$\downarrow$ & SSIM$\downarrow$ & PSNR$\downarrow$ & LPIPS$\uparrow$ \\
        \midrule
        3  & 0.980 $\pm$ 0.017 & 37.45 $\pm$ 5.14 & 0.031 $\pm$ 0.025 & 0.978 & 33.76 & 0.069 \\
        7  & 0.987 $\pm$ 0.011 & 40.60 $\pm$ 5.57 & 0.022 $\pm$ 0.020 & 0.940 & 29.91 & 0.149 \\
        11 & 0.991 $\pm$ 0.010 & 43.83 $\pm$ 7.19 & 0.016 $\pm$ 0.020 & 0.913 & 28.29 & 0.190 \\
        21 & 0.998 $\pm$ 0.001 & 48.21 $\pm$ 5.94 & 0.006 $\pm$ 0.010 & 0.857 & 25.87 & 0.265 \\
        \bottomrule
    \end{tabular}
    }
    \caption{Reconstruction quality and imperceptibility across different Gaussian blur kernel sizes for NeRF-Synthetic scenes. Values represent mean $\pm$ std across scenes.}
    \label{tab:ns_gaussian_blur_results}
\end{table}

\begin{table}
    \centering
    \resizebox{\columnwidth}{!}{
    \begin{tabular}{lccccccc}
        \toprule
        Stddev& \# of scenes & \multicolumn{3}{c}{Poisoned vs Render} & \multicolumn{3}{c}{Poisoned vs Original} \\
        \cmidrule(lr){3-5} \cmidrule(lr){6-8}
         & & SSIM$\uparrow$ & PSNR$\uparrow$ & LPIPS$\downarrow$ & SSIM$\downarrow$ & PSNR$\downarrow$ & LPIPS$\uparrow$ \\
        \midrule
        5  & 8 & 0.851 $\pm$ 0.027 & 31.48 $\pm$ 2.44 & 0.151 $\pm$ 0.029 & 0.691 & 36.57 & 0.105 \\
        10 & 8 & 0.649 $\pm$ 0.029 & 28.57 $\pm$ 1.34 & 0.351 $\pm$ 0.029 & 0.372 & 30.39 & 0.316 \\
        25 & 7 & 0.321 $\pm$ 0.044 & 22.86 $\pm$ 0.93 & 0.599 $\pm$ 0.094 & 0.133 & 22.36 & 0.647 \\
        50 & 4 & 0.260 $\pm$ 0.207 & 17.68 $\pm$ 1.43 & 0.686 $\pm$ 0.346 & 0.070 & 16.49 & 0.895 \\
        \bottomrule
    \end{tabular}
    }
    \caption{Reconstruction quality and imperceptibility across different Gaussian noise standard deviations for NeRF-Synthetic scenes. Values represent mean $\pm$ std across scenes.}
    \label{tab:ns_gaussian_noise_results}
\end{table}

\begin{table}
    \centering
    \resizebox{\columnwidth}{!}{
    \begin{tabular}{lccccccc}
        \toprule
        Transform & \# of scenes & \multicolumn{3}{c}{Poisoned vs Render} & \multicolumn{3}{c}{Poisoned vs Original} \\
        \cmidrule(lr){3-5} \cmidrule(lr){6-8}
         & & SSIM$\uparrow$ & PSNR$\uparrow$ & LPIPS$\downarrow$ & SSIM$\downarrow$ & PSNR$\downarrow$ & LPIPS$\uparrow$ \\
        \midrule
        Rotation ($15^\circ$)          & 8 & 0.974 $\pm$ 0.023 & 34.88 $\pm$ 4.46 & 0.045 $\pm$ 0.033 & 0.732 & 15.33 & 0.288 \\
        Rotation ($30^\circ$)          & 8 & 0.978 $\pm$ 0.021 & 35.82 $\pm$ 4.03 & 0.038 $\pm$ 0.029 & 0.690 & 14.01 & 0.369 \\
        Rotation ($45^\circ$)          & 8 & 0.972 $\pm$ 0.023 & 34.39 $\pm$ 4.22 & 0.046 $\pm$ 0.033 & 0.666 & 13.42 & 0.411 \\
        Random Rotation ($\pm 45^\circ$) & 8 & 0.972 $\pm$ 0.024 & 34.53 $\pm$ 4.39 & 0.047 $\pm$ 0.034 & 0.724 & 15.33 & 0.300 \\
        Shear X ($0.2$)                & 8 & 0.923 $\pm$ 0.061 & 28.25 $\pm$ 7.38 & 0.110 $\pm$ 0.076 & 0.699 & 14.45 & 0.339 \\
        Shear XY ($0.15$)              & 8 & 0.918 $\pm$ 0.059 & 27.58 $\pm$ 7.83 & 0.118 $\pm$ 0.076 & 0.687 & 13.95 & 0.365 \\
        Shear Y ($0.2$)                & 8 & 0.923 $\pm$ 0.055 & 27.61 $\pm$ 6.47 & 0.110 $\pm$ 0.073 & 0.693 & 14.12 & 0.348 \\
        Random Shear XY ($\pm 0.3$)       & 7 & 0.911 $\pm$ 0.046 & 24.65 $\pm$ 4.06 & 0.143 $\pm$ 0.063 & 0.670 & 13.89 & 0.380 \\
        \bottomrule
    \end{tabular}
    }
    \caption{Reconstruction quality and imperceptibility across different geometric transformations for NeRF-Synthetic scenes. Values represent mean $\pm$ std across scenes.}
    \label{tab:ns_geometric_results}
\end{table}

\begin{table}
    \centering
    \resizebox{\columnwidth}{!}{
    \begin{tabular}{lcccccc}
        \toprule
        Quality & \multicolumn{3}{c}{Poisoned vs Render} & \multicolumn{3}{c}{Poisoned vs Original} \\
        \cmidrule(lr){2-4} \cmidrule(lr){5-7}
         & SSIM$\uparrow$ & PSNR$\uparrow$ & LPIPS$\downarrow$ & SSIM$\downarrow$ & PSNR$\downarrow$ & LPIPS$\uparrow$ \\
        \midrule
        10 & 0.958 $\pm$ 0.024 & 33.39 $\pm$ 2.97 & 0.070 $\pm$ 0.056 & 0.920 & 30.08 & 0.093 \\
        25 & 0.959 $\pm$ 0.023 & 32.70 $\pm$ 2.61 & 0.058 $\pm$ 0.032 & 0.945 & 32.61 & 0.046 \\
        50 & 0.961 $\pm$ 0.025 & 32.86 $\pm$ 3.52 & 0.054 $\pm$ 0.033 & 0.958 & 34.63 & 0.025 \\
        75 & 0.963 $\pm$ 0.026 & 32.26 $\pm$ 2.97 & 0.055 $\pm$ 0.037 & 0.969 & 36.96 & 0.012 \\
        \bottomrule
    \end{tabular}
    }
    \caption{Reconstruction quality and imperceptibility across different JPEG compression quality levels for NeRF-Synthetic scenes. Values represent mean $\pm$ std across scenes.}
    \label{tab:ns_jpeg_compression_results}
\end{table}

\begin{figure}
    \centering
    \includegraphics[width=0.78\linewidth]{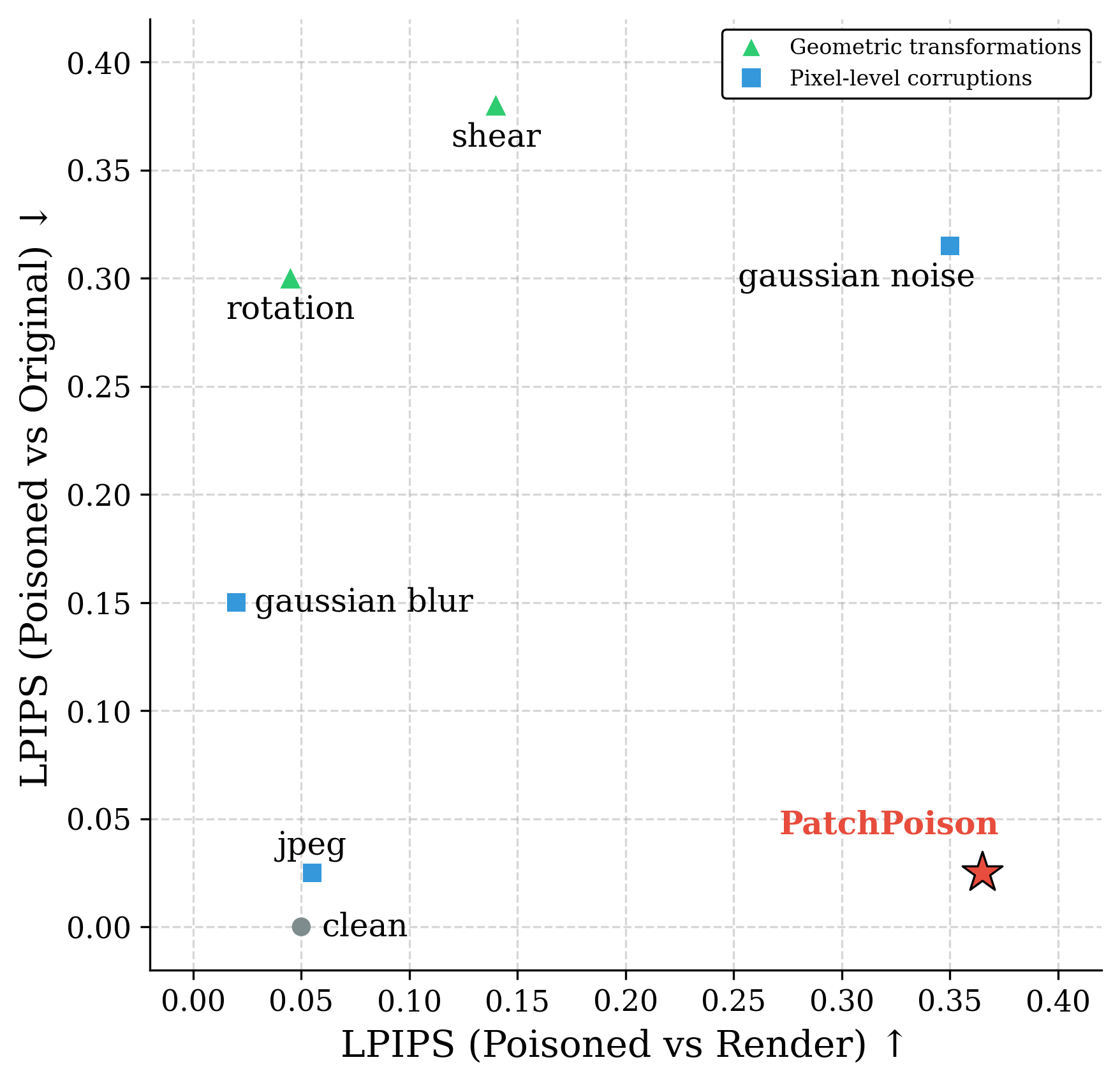}
    \caption{\textbf{Trade-off between reconstruction degradation and imperceptibility.} An effective method should lie in the bottom-right region. PatchPoison achieves this optimal balance, while baselines either fail to degrade reconstruction or introduce visible artifacts.}
    \label{fig:baseline_vs_pp}
\end{figure}

In contrast, PatchPoison uniquely achieves strong reconstruction degradation while maintaining high visual similarity, as illustrated in Fig.~\ref{fig:baseline_vs_pp}.

\subsection{Generalization to Real-World Scenes}

\begin{figure}
    \centering
    \includegraphics[width=\linewidth]{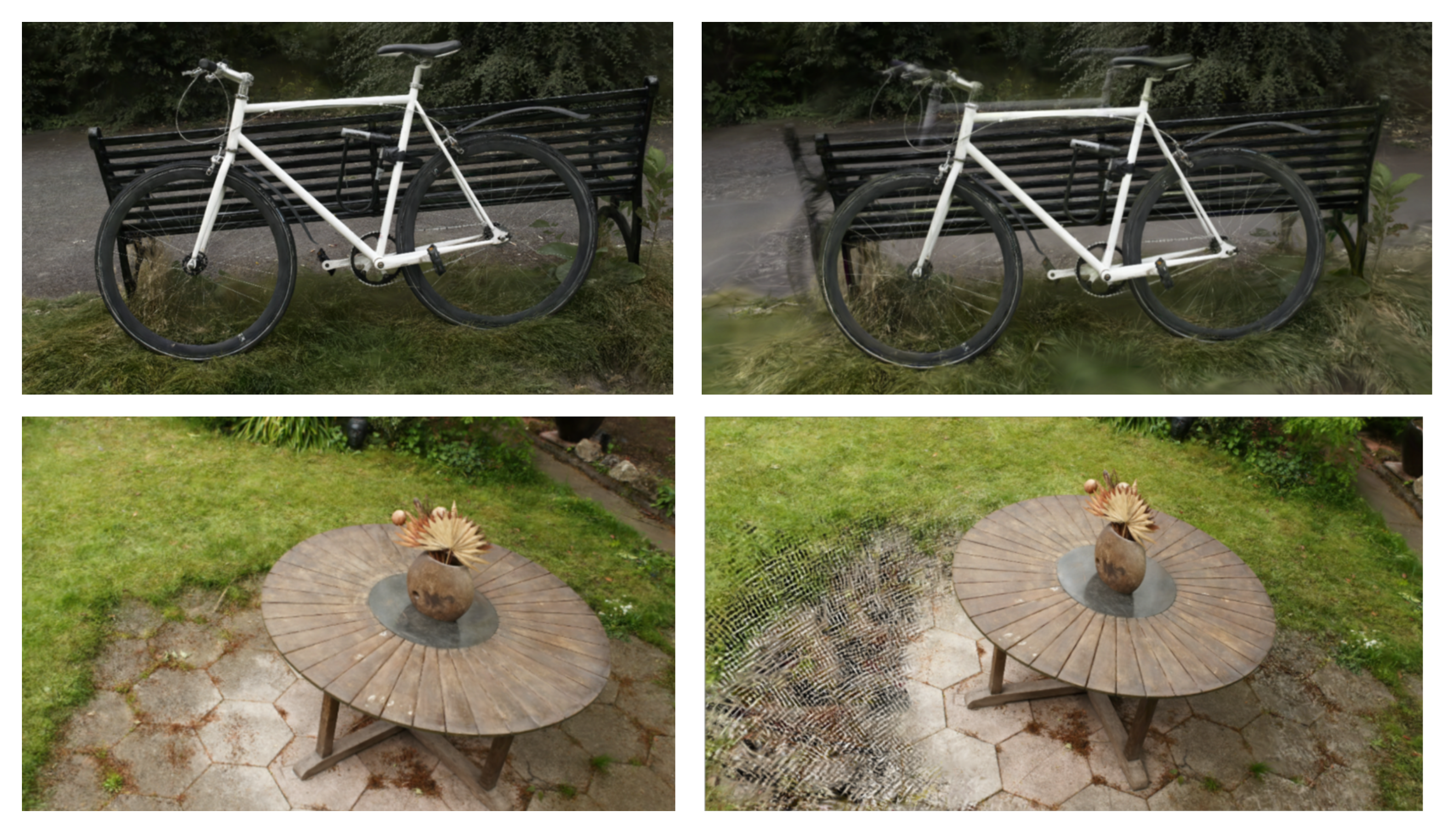}
    \caption{\textbf{Results on Mip-NeRF 360 scenes.} In structured scenes such as \textit{bicycle}, PatchPoison introduces visible distortions. In complex natural scenes such as \textit{garden}, the patch blends into the background, reducing visual artifacts while still increasing reconstruction error.}
    \label{fig:mip-360-fails}
\end{figure}

Table~\ref{tab:mip-360-patch size} evaluates PatchPoison on the Mip-NeRF~360 dataset. While quantitative metrics indicate degradation similar to synthetic scenes, visual inspection reveals that the patch often blends into natural backgrounds, reducing its effectiveness. Figure~\ref{fig:mip-360-fails} shows our results on Mip-NeRF~360.

\begin{table}
    \centering
    \resizebox{\columnwidth}{!}{
    \begin{tabular}{lcccccc}
        \toprule
        Patch & \multicolumn{3}{c}{Poisoned vs Render} & \multicolumn{3}{c}{GT vs Poisoned} \\
        \cmidrule(lr){2-4} \cmidrule(lr){5-7}
         & SSIM$\uparrow$ & PSNR$\uparrow$ & LPIPS$\downarrow$ & SSIM$\downarrow$ & PSNR$\downarrow$ & LPIPS$\uparrow$ \\
        \midrule
        2\%    & 0.923 $\pm$ 0.035 & 30.57 $\pm$ 1.66 & 0.119 $\pm$ 0.039 & 0.982 & 36.37 & 0.008 \\
        6\%    & 0.878 $\pm$ 0.104 & 25.75 $\pm$ 2.05 & 0.164 $\pm$ 0.103 & 0.979 & 27.44 & 0.014 \\
        12.5\% & 0.876 $\pm$ 0.076 & 21.50 $\pm$ 0.99 & 0.161 $\pm$ 0.069 & 0.961 & 21.10 & 0.034 \\
        25\%   & 0.752 $\pm$ 0.204 & 15.58 $\pm$ 1.88 & 0.309 $\pm$ 0.286 & 0.893 & 15.07 & 0.110 \\
        \bottomrule
    \end{tabular}
    }
    \caption{Reconstruction quality and imperceptibility aggregated across scenes for different patch sizes in terms of \% of the width of the image. Values represent mean $\pm$ std across scenes in the MIP-360 dataset.}
    \label{tab:mip-360-patch size}
\end{table}

\paragraph{Limitation.}
PatchPoison is less effective in complex real-world scenes with rich textures, where the patch becomes less distinguishable from the background (Fig.~\ref{fig:mip-360-fails}). Learning-based SfM methods (DUSt3R~\cite{wang2024dust3r}, MASt3R~\cite{leroy2024mast3r}, MUST3R~\cite{must3r_cvpr25}) that avoid explicit keypoint matching may be more resilient. This suggests that future work should explore adaptive or content-aware poisoning strategies.

\paragraph{Ethical considerations.}
PatchPoison is intended as a tool for content owners to protect multi-view imagery from unauthorized 3D reconstruction. We do not condone its misuse in harmful or deceptive contexts. The release of code will be accompanied by an access policy consistent with responsible disclosure norms.
\section{Conclusion}
\label{sec:conclusion}

We have presented PatchPoison, a simple and effective method for poisoning multi-view
image datasets to prevent unauthorized 3D reconstruction with NeRF and 3D
Gaussian Splatting.
By injecting a small high-frequency checkerboard patch at a fixed corner of each
image, PatchPoison corrupts the feature matching stage of COLMAP, leading to
misaligned camera poses and severely degraded reconstructions.
Crucially, the poisoned images remain visually indistinguishable from the originals
by human observers.
Extensive experiments across 8 scenes demonstrate consistent effectiveness
with a patch as small as $12\times12$ pixels less than $0.03\%$ of the image
area while maintaining poisoned-image fidelity above $\text{SSIM} = 0.999$.
PatchPoison operates as a lightweight post-processing step requiring no access to the
reconstruction pipeline, making it immediately deployable by content creators.
{
    \small
    \bibliographystyle{ieeenat_fullname}
    \balance
    \bibliography{main}

@String(CVPR= {IEEE Conf. Comput. Vis. Pattern Recog.})

@String(ECCV= {Eur. Conf. Comput. Vis.})

@String(CVPR  = {CVPR})

@String(ECCV  = {ECCV})

@inproceedings{mildenhall2020nerf,
  title     = {{NeRF}: Representing Scenes as Neural Radiance Fields for View Synthesis},
  author    = {Mildenhall, Ben and Srinivasan, Pratul P. and Tancik, Matthew and
               Barron, Jonathan T. and Ramamoorthi, Ravi and Ng, Ren},
  booktitle = {European Conference on Computer Vision (ECCV)},
  year      = {2020}
}

@article{kerbl3Dgaussians,
  title   = {{3D Gaussian Splatting} for Real-Time Radiance Field Rendering},
  author  = {Kerbl, Bernhard and Kopanas, Georgios and Leimk{\"u}hler, Thomas and
             Drettakis, George},
  journal = {ACM Transactions on Graphics},
  volume  = {42},
  number  = {4},
  year    = {2023}
}

@inproceedings{schoenberger2016sfm,
  title     = {Structure-from-Motion Revisited},
  author    = {Sch{\"o}nberger, Johannes Lutz and Frahm, Jan-Michael},
  booktitle = {IEEE/CVF Conference on Computer Vision and Pattern Recognition (CVPR)},
  year      = {2016}
}

@inproceedings{barron2022mipnerf360,
  title     = {Mip-{NeRF} 360: Unbounded Anti-Aliased Neural Radiance Fields},
  author    = {Barron, Jonathan T. and Mildenhall, Ben and Verbin, Dor and
               Srinivasan, Pratul P. and Hedman, Peter},
  booktitle = {IEEE/CVF Conference on Computer Vision and Pattern Recognition (CVPR)},
  year      = {2022}
}

@inproceedings{wang2024dust3r,
  title     = {{DUSt3R}: Geometric 3D Vision Made Easy},
  author    = {Wang, Shuzhe and Leroy, Vincent and Cabon, Yohann and
               Chidlovskii, Boris and Revaud, Jerome},
  booktitle = {IEEE/CVF Conference on Computer Vision and Pattern Recognition (CVPR)},
  year      = {2024}
}

@article{leroy2024mast3r,
  title   = {{MASt3R}: Matching and Stereo 3D Reconstruction},
  author  = {Leroy, Vincent and Cabon, Yohann and Revaud, Jerome},
  journal = {arXiv preprint arXiv:2406.09756},
  year    = {2024}
}

@inproceedings{
    lu2025poisonsplat,
    title={Poison-splat: Computation Cost Attack on 3D Gaussian Splatting},
    author={Jiahao Lu and Yifan Zhang and Qiuhong Shen and Xinchao Wang and Shuicheng YAN},
    booktitle={The Thirteenth International Conference on Learning Representations},
    year={2025},
    url={https://openreview.net/forum?id=ExrEw8cVlU}
}

@article{hong2025gausstrap,
  title={GaussTrap: Stealthy Poisoning Attacks on 3D Gaussian Splatting for Targeted Scene Confusion},
  author={Hong, Jiaxin and Chen, Sixu and Sun, Shuoyang and Yu, Hongyao and Fang, Hao and Tan, Yuqi and Chen, Bin and Qi, Shuhan and Li, Jiawei},
  journal={arXiv preprint arXiv:2504.20829},
  year={2025}
}

@article{ding2025kba,
    title   = {Kaleidoscopic Background Attack: Disrupting Pose Estimation with Multi-Fold Radial Symmetry Textures},
    author  = {Xinlong Ding and Hongwei Yu and Jiawei Li and Feifan Li and Yu Shang and Bochao Zou and Huimin Ma and Jiansheng Chen},
    journal = {arXiv preprint arXiv:2507.10265},
    year    = {2025}
}

@article{jiang2024ipa,
  title={Ipa-nerf: Illusory poisoning attack against neural radiance fields},
  author={Jiang, Wenxiang and Zhang, Hanwei and Zhao, Shuo and Guo, Zhongwen and Wang, Hao},
  journal={arXiv preprint arXiv:2407.11921},
  year={2024}
}

@article{zeybey2024gaussian,
  title={Gaussian splatting under attack: Investigating adversarial noise in 3d objects},
  author={Zeybey, Abdurrahman and Ergezer, Mehmet and Nguyen, Tommy},
  journal={arXiv preprint arXiv:2412.02803},
  year={2024}
}

@article{zhang2024gs,
  title={Gs-hider: Hiding messages into 3d gaussian splatting},
  author={Zhang, Xuanyu and Meng, Jiarui and Li, Runyi and Xu, Zhipei and Zhang, Yongbing and Zhang, Jian},
  journal={Advances in Neural Information Processing Systems},
  volume={37},
  pages={49780--49805},
  year={2024}
}

@article{li2024gaussianstego,
  title={Gaussianstego: A generalizable stenography pipeline for generative 3d gaussians splatting},
  author={Li, Chenxin and Liu, Hengyu and Fan, Zhiwen and Li, Wuyang and Liu, Yifan and Pan, Panwang and Yuan, Yixuan},
  journal={arXiv preprint arXiv:2407.01301},
  year={2024}
}

@article{tan2024water,
  title={Water-gs: Toward copyright protection for 3d gaussian splatting via universal watermarking},
  author={Tan, Yuqi and Liu, Xiang and Xie, Shuzhao and Chen, Bin and Xia, Shu-Tao and Wang, Zhi},
  journal={arXiv preprint arXiv:2412.05695},
  year={2024}
}

@article{huang2024gaussianmarker,
  title={Gaussianmarker: Uncertainty-aware copyright protection of 3d gaussian splatting},
  author={Huang, Xiufeng and Li, Ruiqi and Cheung, Yiu-ming and Cheung, Ka Chun and See, Simon and Wan, Renjie},
  journal={Advances in Neural Information Processing Systems},
  volume={37},
  pages={33037--33060},
  year={2024}
}

@inproceedings{chen2025guardsplat,
  title={GuardSplat: efficient and robust watermarking for 3d gaussian splatting},
  author={Chen, Zixuan and Wang, Guangcong and Zhu, Jiahao and Lai, Jianhuang and Xie, Xiaohua},
  booktitle={Proceedings of the Computer Vision and Pattern Recognition Conference},
  pages={16325--16335},
  year={2025}
}

@inproceedings{must3r_cvpr25,
      title={MUSt3R: Multi-view Network for Stereo 3D Reconstruction}, 
      author={Yohann Cabon and Lucas Stoffl and Leonid Antsfeld and Gabriela Csurka and Boris Chidlovskii and Jerome Revaud and Vincent Leroy},
      booktitle = {CVPR},
      year = {2025}
}

@article{snavely2006photo,
author = {Snavely, Noah and Seitz, Steven M. and Szeliski, Richard},
title = {Photo tourism: exploring photo collections in 3D},
year = {2006},
issue_date = {July 2006},
publisher = {Association for Computing Machinery},
address = {New York, NY, USA},
volume = {25},
number = {3},
issn = {0730-0301},
url = {https://doi.org/10.1145/1141911.1141964},
doi = {10.1145/1141911.1141964},
journal = {ACM Trans. Graph.},
month = jul,
pages = {835–846},
numpages = {12}
}

@article{song2024geometry,
  title={Geometry cloak: Preventing tgs-based 3d reconstruction from copyrighted images},
  author={Song, Qi and Luo, Ziyuan and Cheung, Ka Chun and See, Simon and Wan, Renjie},
  journal={Advances in Neural Information Processing Systems},
  volume={37},
  pages={119361--119385},
  year={2024}
}
    \balance
}
\end{document}